# Training Deep Learning Models with Hybrid Datasets for Robust Automatic Target Detection on real SAR images

Benjamin Camus
Scalian DS
2 rue Antoine Becquerel
35700 Rennes, France
benjamin.camus@scalian.com

Théo Voillemin
Scalian DS
2 rue Antoine Becquerel
35700 Rennes, France
theo.voillemin@scalian.com

Corentin Le Barbu
Scalian DS
2 rue Antoine Becquerel
35700 Rennes, France
corentin.lebarbu@scalian.com

Jean-Christophe Louvigné
DGA Maîtrise de l'Information
BP 7
35998 Rennes CEDEX 9
jean-christophe.louvigne@intradef.gouv.fr

Carole Belloni
DGA Maîtrise de l'Information
BP 7
35998 Rennes CEDEX 9
carole.belloni@intradef.gouv.fr

Emmanuel Vallée
DGA Maîtrise de l'Information
BP 7
35998 Rennes CEDEX 9
emmanuel.vallee@intradef.gouv.fr

*Abstract* - **In this work, we propose to tackle several challenges hindering the development of Automatic Target Detection (ATD) algorithms for ground targets in SAR images. To address the lack of representative training data, we propose a Deep Learning approach to train ATD models with synthetic target signatures produced with the MOCEM simulator. We define an incrustation pipeline to incorporate synthetic targets into real backgrounds. Using this hybrid dataset, we train ATD models specifically tailored to bridge the domain gap between synthetic and real data. Our approach notably relies on massive physics-based data augmentation techniques and Adversarial Training of two deep-learning detection architectures. We then test these models on several datasets, including (1) patchworks of real SAR images, (2) images with the incrustation of real targets in real backgrounds, and (3) images with the incrustation of synthetic background objects in real backgrounds. Results show that the produced hybrid datasets are exempt from image overlay bias. Our approach can reach up to 90% of Average Precision on real data while exclusively using synthetic targets for training.**

## I. Introduction

Automatic Target Detection (ATD) on SAR images consists in automatically localizing objects of interest (usually vehicles) in large background scenes imaged by a radar. Several works have demonstrated that Deep Learning algorithms can tackle this challenge with good results [1, 2]. However, the training was performed on real measurements. The problem is that acquiring sufficient labelled data of objects of interests may be very difficult, if not impossible, in many practical applications. The solution to overcome this issue is to produce synthetic datasets with a SAR simulator. Indeed, simulation enable the production of large and diversified datasets in a timely manner, and to automatically label the data. However, synthetic data are not as representative as true measurements, because (1) simulation is intrinsically grounded on simplifying assumptions (i.e. models), (2) the knowledge of the objects effectively deployed is necessarily partial, and (3) objects of interest are highly variable. In Machine Learning terms, this mean that the synthetic and real data distribution differs, which correspond to the well-known Dataset Shift problem [3]. The challenge is then to build ATD models that can generalize from synthetic data to real measurements.

Previous works have addressed this challenge in a related task: SAR Automatic Target Recognition (ATR) [4, 5]. ATR consists in automatically classifying objects of interest from small SAR vignettes. It can be performed after ATD to identify the detected vehicles. In [6], we proposed an original approach to train ATR models with synthetic SAR images. Our approach, called ADASCA, combined (1) a set of Deep Learning techniques to improve the generalization ability of ATR models, with (2) massive physic-based data-augmentation performed on-the-fly during training to improve the representativeness of synthetic datasets. We have demonstrated that models trained exclusively on synthetic data using ADASCA reach an accuracy of 75 % on real measurements of the MSTAR reference dataset [7]. This represents an increase of 52 % compared to standard Deep Learning classification algorithms, and 20 % compared to the best SAR ATR algorithms of the literature.

In this work, we propose to adapt ADASCA to the ATD task to train models with synthetic data. We modify two state-of-the-art detection algorithms (Faster-RCNN [8] and RetinaNet [9]) to enhance their generalization ability when training with synthetic data and testing on real data.

Importantly, we also design a physic-based massive data generation pipeline to produce the training and testing data for the ATD models. This pipeline performs data augmentation on synthetic SAR signatures and insert them at random position in several background clutter images. The data-augmentation is performed on-the-fly directly during training of the models. To perform our simulations, we use the MOCEM software developed by SCALIAN DS for DGA-MI (i.e. the French MoD) [10]. We use MSTAR clutter images as background. Thus, we train our models with real background measurements and synthetic target signatures.

To test our models, we produce various ATD datasets. We use the publicly available MSTAR datasets that include (1) small 128x128 vignettes with a single target per image for ATR, and (2) large background clutter images without target but that can include background objects. We propose two new methods to build large backgrounds with targets for ATD. The first one, inspired by [1], consists of building a patchwork with the MSTAR vignettes. The second one consists of using a segmentation model to separate the targets and their shadows in the MSTAR vignettes. The measured targets and shadows are then overlayed in the background clutter images. To this end, we adapt ADASCA to train a Deep Learning segmentation model with synthetic data. We show then that the model generalizes well on the real MSTAR measurements. This a significant secondary contribution of this paper.

The rest of the paper is as follows. In section II.A we detail the production of the synthetic training data with the MOCEM simulator. In Section II.B, we describe our training dataset production pipeline. In Section II.C, we detail the test datasets used to evaluate that our ATD models trained with synthetic data can generalize to real measurements. Section II.D discusses the dataset train/test split. In Section II.E, we present the training algorithms used. Section II.F introduces the evaluation metrics. Finally, in section III.A, we show that we obtain good evaluation results.

## II. MATERIAL AND METHODS

### A. Simulation of targets SAR signatures

To train our ATD models, we produce a synthetic version of the SAR images of the MSTAR (Moving and Stationary Target Acquisition and Recognition) public targets. As shown in Figure 1, MSTAR is comprised of airborne SAR images taken at different incidence and azimuth angles for ten classes of military ground vehicles: 2S1, BMP2, BRDM2, BTR60, BTR70, D7, T62, T72, ZIL131 and ZSU23-4.

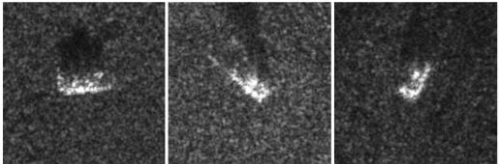

*Figure 1. Samples of MSTAR vignettes of the BRDM2 (left), BTR60 (center) and BTR70 (right).*

We use the MOCEM simulator to produce our synthetic data. MOCEM is a CAD-based simulator that perform geometrical analysis of 3D models of the object of interest to search for canonical radar effects (i.e. trihedral, dihedral). We take off-the-shelf CAD models of the vehicles available on Internet and simplify the meshes (when appropriate) to speed-up the simulation time. We associate generic electro-magnetic (EM) materials (i.e. with well-defined reflectivity, roughness and dielectric constant) to the different facets. Finally, we configure MOCEM to simulate the MSTAR sensor function (i.e. with similar range/cross-range sampling and resolution, thermal noise level, and window function).

Following our previous work [6], we do not try to fine-tune the CAD models to precisely match the ground truth of the MSTAR measurement. Thus, like in a true operational context, the CAD models used to produce the synthetic data may differ from the actual vehicles imaged on the measurements. For instance, the orientations of the articulated parts (e.g. turret and gun), the outside equipment (e.g. fuel barrel), and the EM reflectivity of the materials may be different between training and test. These constraints are close than the ones we may find in a real operational scenario.

For each vehicle, we ran parametric simulation to produce images at every 0.5° of azimuth for the 15°, 16° and 17° depression angles. For each angle, we produce (1) a complex image containing only the signature of the target, and (2) a shadow mask indicating pixels that should contains the contributions of the background scatterers (see Figure 2). Hence, our final dataset comprises a total of 21600 pairs of signatures and shadow masks.

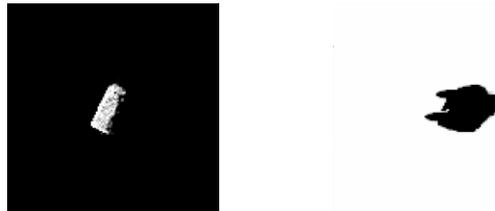

*Figure 2. Sample of data generated with MOCEM: target signature (left) shadow mask (right).*

### B. Train dataset production pipeline

We propose the following dataset production pipeline to train ATD models with synthetic data.

Firstly, we developed a pipeline for producing synthetic training data, based on our previous work for the ATR problem [6]. This pipeline generates simulated data on-the-fly and in real time to train a detection algorithm. It takes as input the synthetic target and shadow mask pairs produced in Section II.A. together with the MSTAR backgrounds [7].

The first phase consists of selecting randomly a background and one to three targets and shadow mask pairs. The overlaying phase manages the non-overlaping of the targets, it also precalculates the bounding boxes of each targets for auto labelling. Note that we made the decision to overlay the targets anywhere in the background, whether on obstacles (buildings, forests, etc.) or shadows. We believe that training the neural network on target overlays that are not always representative of real cases makes it more robust. We use domain randomization strategy [12] to introduce as much variations in the synthetic data as possible. For each synthetic data to overlay, a set of different random parameters is selected to apply the following randomizations: cross-range and range resolution, the thermal noise of the sensor, the number of targets and their overlayed positions. We also apply a random dropout on half of the brightest points on each target to introduce variations in their signatures. MSTAR backgrounds are randomly cropped into 640x640 sub-images.

Once all these parameters chosen, the image overlay pipeline is operated, with the following steps (Figure 3):
1. The shadow masks are used to cut the background out.
2. A random complex thermal noise is added on the background, and the sensor function is applied to blur the background into the overlayed shadows.
3. The shadow mask is reused to extract the noisy target shadows produced at operations 2 and overlayed them into the original background.
4. After application of the sensor function, the simulated targets are added on the results of the previous operation.
5. Finally, a quarter power magnitude LUT is applied to get an 8bits output image that can be fed to an ATD algorithm.

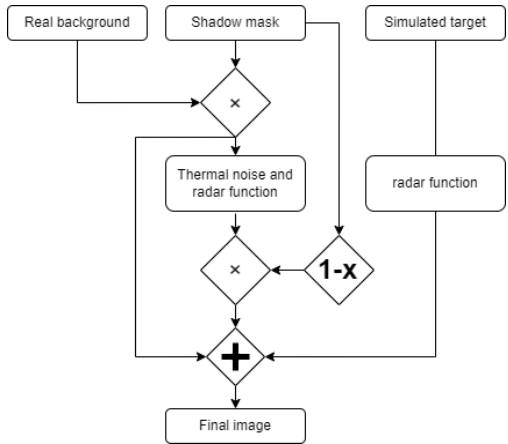

Figure 3. Worklow of our train data production pipeline

The entire pipeline is implemented in Tensorflow [13], allowing data generation and augmentation in real time, taking advantage of the parallelization of GPU architectures. We present an output of our pipeline in Figure 4.

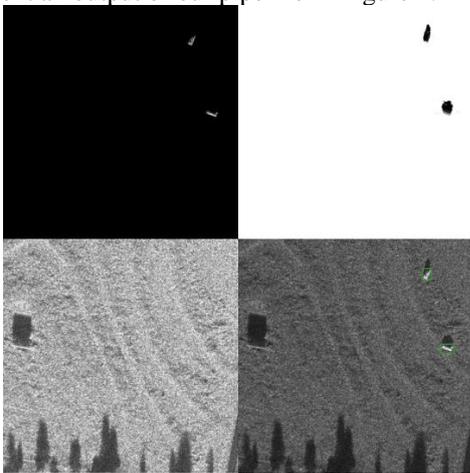

Figure 4. Example of a data produced by our automatic overlay pipeline. Inputs: targets (top-left), supports (top-right), MSTAR background (bottom-left). Output: Image produced (bottom-right) also displaying the precalculated bounding boxes.

### C. Test datasets production pipelines

Using the MSTAR targets and background images we produced four test datasets to evaluate our ATD models trained with synthetic targets (as per Section II.B) in the following four test configurations:

*1) **Real targets in real backgrounds** :* To overlay real target images into a background image, it is necessary to first learn a segmentation model to separate the target, shadow and background in the 128x128 synthetic vignettes generated by MOCEM in Section II.A. The segmentation labels are directly computed from the target signature and shadow mask. We used a nnU-Net algorithm [18], and our ADASCA data-augmentation pipeline to train the model with SAR-based domain randomization [6]. After training, the model is used to predict segmentation mask on the real MSTAR vignettes. As shown in *Figure 5*, we observe very good segmentation on the measured data.

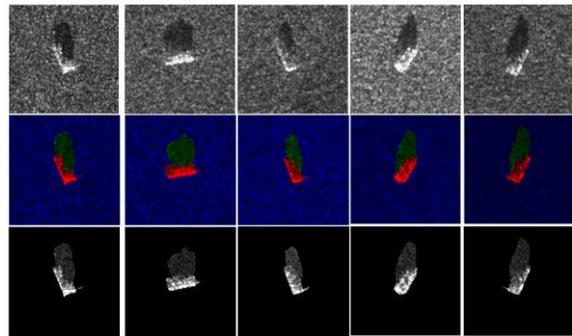

Figure 5. Example of measured target and shadow extraction thanks to our segmentation model. Top: original MSTAR images. Center: segmentation of the targets (red) shadows (green) and backgrounds (blue). Bottom: images with the background clutter removed.

Using the segmentation results, we extracted the targets and shadows in the MSTAR vignettes and overlay them in the MSTAR background images thanks to our automatic pipeline of Section 0. The bounding boxes labels are directly deduced from the target segmentation masks. An example of resulting image is shown in Figure 6.

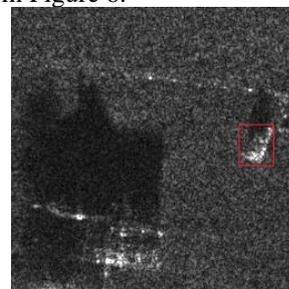

Figure 6 - Example of a measured target (middle right) extracted by our segmentation algorithm and overlayed in a measured background.

*2) **Synthetic targets in real backgrounds**:* To build this dataset, we used our data generation pipeline presented in Section 0. It enables us to establish baseline performance of the ATD model on synthetic targets.

*3) **Patchwork of real vignettes**:* For this dataset, we form large background by concatenating the MSTAR vignettes to form 4x4 patchworks. To ensure smooth transitions, we overlapped the vignettes on 16 pixels strips both on x and y. In the overlapping zones, we applied a linear weighting of the images amplitude and sum the overlapping pixels. This introduce a fade between the vignettes. Like with dataset 1, the bounding box labels are deduced from the target segmentation masks. We produced 200 of these patchwork by selecting random MSTAR vignettes (*Figure 7*).

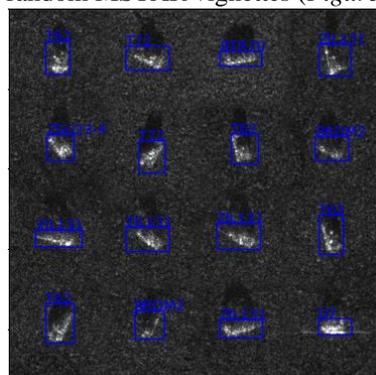

Figure 7. Example of patchwork and labels produced for dataset 2.

Contrarily to other datasets, the background does not contain any object. Hence, the detection task should be easier. However, we are sure that there is no segmentation and image overlay bias because the targets remain in their original background clutter. Thus, this dataset serves us to demonstrate that our ATD models do not exploit some potential image overlay bias to easily detect the targets.

*4) Synthetic background objects in real backgrounds*: Here we simulate synthetic signatures of background objects (i.e. a tree and a country house) and overlay them in the MSTAR background. If our ATD models exploit potential overlay bias to detect the target, then it should also detect these objects, although they do not correspond to our objects of interest. Hence, by demonstrating that this is not the case, we will demonstrate that our ATD models are robust to Out Of Distribution data., like with dataset 3. To produce dataset 4, we took on-the-shelf CAD models of a tree and a country house. As for the vehicles of Section II.A, we simplified the meshes, and associate generic EM materials to the facets. We produced images every 5° in an angular sector of [0°; 100°] at a depression of 15°. The resulting 21 images of each object are then overlayed in the MSTAR background using our pipeline of Section II.B (*Figure 8*).

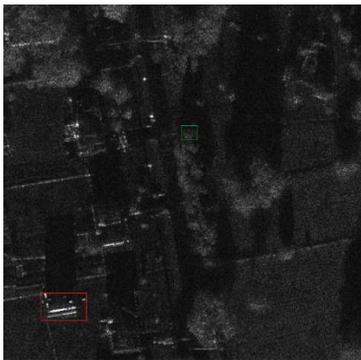

*Figure 8 - Example of simulated data with a simulated house (red) and a simulated tree (green)*

### D. Data management

It is important to note that, we do not use the same MSTAR background images for training and test. The ATD models are then evaluated on completely new data never seen during the training phase. The two datasets were split randomly so that each of the ten targets were represented in both sets, but certain azimuths were not present the test dataset. As for the MSTAR backgrounds, the split was also done randomly but a subsequent visual verification was made so that the urban and field contexts were present in both datasets. We also split the synthetic target and shadow mask pairs so that those overlayed during the testing phase were never used by the network during its training (Table 1).

|  | Train | Test |
|---|---|---|
| MSTAR background | 160 | 20 |
| Target shadow pairs | 19 440 | 2160 |

*Table 1 - Numbers of MSTAR background and synthetic targets and shadows pairs in our training and testing dataset*

### E. Automatic Target Detection architecture and training

We base our experimented on two state-of-the-art detection algorithms, Faster-RCNN [8] and RetinaNet [9].

Faster-RCNN is a two-stage object detection network with a first phase computing features of region proposal and a second phase using these characteristics in order to propose detection for each region. Our implementation use a pretrained VGG-16 network [14] as backbone using Tensorflow 2 / Keras frameworks.

RetinaNet is a one-stage object detection algorithm, doing detection and classification in a single pass. For this network, we use a pretrained ResNet50 network [15] as backbone using PyTorch framework.

For both algorithms, we use a Stochastic Gradient Descent optimizer with a momentum of 0.9 and a weight decay rate of 5e-4. Following our previous work [6], to increase model generalizability we also implemented adversial training [16] on both networks using the Fast Gradient Sign Method [17]. Both networks have similar outputs. For an input image, both networks return a set of predictions, each comprising a confidence score and the predicted bounding box coordinates. Figure 9 presents the full pipeline of the training process of our ATD approach.

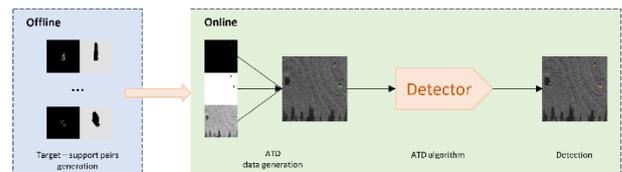

*Figure 9 - Full training pipeline. A first offline stage is precalculated to generate all target support pairs. A second online stage is a real-time pipeline that generates training object detection data that is used to train an automatic target detection neural network.*

Using our data generation pipeline of Section II.B we trained our ATD models with the synthetic target data of Section II.A, and the real background measurement of MSTAR. Please note that for our study, we are considering a pure detection problem, meaning that the network does not classify the targets.

### F. Metrics

For each experiment, we used the usual metrics for evaluating detection, namely: the Intersection over Union (IoU), Precision x Recall curve and the Average Precision (AP).
IoU is used to determine whether a detector prediction should be considered valid or not. Each prediction is compared to all ground truths (GT) that are not already paired with another prediction. The prediction is valid if the intersection of the two bounding box surfaces divided by their union exceeds an arbitrarily chosen threshold. Thus, we can determine the number of True Positive (TP) when a prediction match a GT, False Negative (FN) when a prediction does not match any GT and False Positive (FP) when a GT is not detected. The precision and recall metrics are computed in a standard way:

$$Precision = \frac{TP}{TP + FP} = \frac{Good\ detections}{All\ detections}$$

$$Recall = \frac{TP}{TP + FN} = \frac{Good\ detections}{Number\ of\ GT}$$

The Precision x Recall curve is computed by fixing an IoU threshold and varying the confidence threshold (the threshold that the confidence score of a prediction must exceed to be considered). For each confidence threshold, the precision and the recall values are calculated and plotted. Finally, we compute the Average Precision by measuring the area under the Precision x Recall curve.

Since our targets and their bounding boxes are relatively small compared to other state-of-the-art sensing contexts, small variations result in significant IoU changes. Thus, we noticed that visually valid predictions are not valid with respect to the IoU threshold of 0.5 usually used in the state of the art. Therefore, we also present the AP obtained with a IoU threshold of 0.25 together with Precision x Recall curves and the AP as a function of different thresholds.

## III. RESULTS

In the following, we detail the results obtained with various experiments performed to validate our ATD models trained with synthetic SAR target signatures and real background measurements.

### A. Results on the four test datasets

*1) Synthetic targets in real backgrounds* : we observe that in both cases, the predicted bounding boxes are visually valid with respect to their ground truth. However, Faster-CNN only reaches a score of 36.4% when using an IoU threshold of 50%. This is caused by the IoU metric that is too sensitive with small bounding boxes. That is why we lowered the IoU threshold to 25 %. In this case, the accuracy of Faster-RCNN becomes 83.4%. RetinaNet still seems to fit better the bounding boxes dimensions achieving a good result of 83.5 % even with an IoU threshold of 50%, and 88 % with an IoU threshold of 25% (Table 2).

|           | Faster-RCNN | RetinaNet |
|-----------|-------------|-----------|
| $AP_{50}$ | 36.4%       | **83.5%** |
| $AP_{25}$ | 83.4%       | **88.1%** |

*Table 2- Average Precision on different IoU threshold (50% and 25%) of Faster-RCNN and RetinaNet over the simulated targets on measured backgrounds dataset*

*2) Real targets in real backgrounds:* We observe similar results than those already described in Section III.A.1) regarding the IoU threshold (Figure 10). However, it should be noted that, for the two neural networks, the detection results (89.3% for Faster-RCNN and 91.4% for RetinaNet) are significantly higher with real test targets than with synthetic ones even though we trained our models with synthetic data (Table 3). This demonstrates that the simulation of our targets and the data augmentation operations carried out are sufficiently powerful so that the ATD models can generalize and operate from synthetic data to unobserved measured data.

|           | Faster-RCNN | RetinaNet |
|-----------|-------------|-----------|
| $AP_{50}$ | 49.2%       | **83.2%** |
| $AP_{25}$ | 89.3%       | **91.4%** |

*Table 3 - Average Precision on different IoU threshold (50% and 25%) of Faster-RCNN and RetinaNet over the measured targets on measured backgrounds dataset*

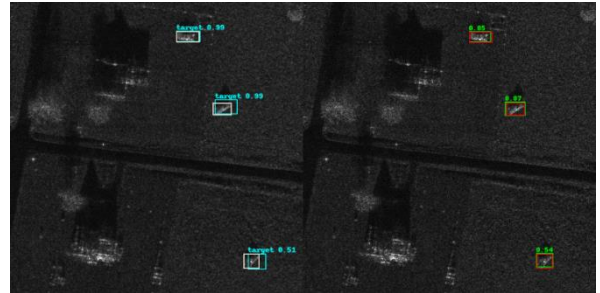

*Figure 10 - Example of detections on the measured dataset. Left: Faster-RCNN (blue: prediction, white: GT). Right: RetinaNet (green: prediction, red: GT).*

*3) Patchwork of real vignettes:* In terms of scene context, this dataset is simpler than the previous ones since there is no object (trees, buildings, etc.) other than the targets in the images. However, as the targets remains in their original clutter, this dataset has the particularity of not including any overlaying operations. This allows us to bypass a potentiel bias where the ATD model might simply learn to detect some potential overlaying flaws rather than identifying targets. (Figure 11).

|           | Faster-RCNN | RetinaNet |
|-----------|-------------|-----------|
| $AP_{50}$ | 54.4%       | **73.3 %** |
| $AP_{25}$ | **98.3%**   | 85.1%     |

*Table 4 - Average Precision on different IoU threshold (50% and 25%) of Faster-RCNN and RetinaNet over the measured patchworks dataset*

We still obtain similar or even better results than with the two previous datasets (Table 4), although both are closer to the dataset used in training. This demonstrates that our approach does not create bias during learning with respect to the overlaying operation.

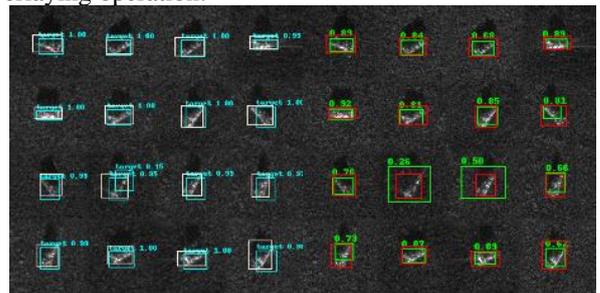

*Figure 11 - Example of detections on the patchwork dataset. Left: Faster-RCNN (blue: prediction, white: GT). Right: RetinaNet (green: prediction, red: GT).*

*4) Synthetic background objects in real background:* In this experiment, given that there is no target vehicle, and that we do not want our model to detect the overlayed background objects, the objective is to have a poor average precision. Here again, the two neural networks obtain very satisfactory results. Considering the same metrics of average precision and IoU threshold, less than 10% of overlayed background objects are detected as targets (Table 5). The rare detection are always the bright points of the simulated houses.

|  | Faster-RCNN | RetinaNet |
|---|---|---|
| $AP_{50}$ | **0.2%** | **0.01%** |
| $AP_{25}$ | 9.8% | 7.9% |

*Table 5 - Average Precision on different IoU threshold (50% and 25%) of Faster-RCNN and RetinaNet over the simulated background objects on measured backgrounds dataset*

### B. Global Performance Analysis

The RetinaNet network tends to be more flexible on its detections than Faster-RCNN, leading to a greater number of false detections but also to better detection of GT. We also observe, for the two networks and for the four test datasets, the same behavior with respect to the sensitivity of the IoU threshold. In each case, the networks offer very good AP results on each of the datasets for an IoU threshold below 30%. Beyond this threshold, we very clearly notice a collapse of the results, the predicted bounding boxes being no longer considered valid (Figure 12). It is also interesting to note that for the two networks, there is no overfitting on the simulated targets given that we obtain similar or even better results on the measured dataset (Figure 13).

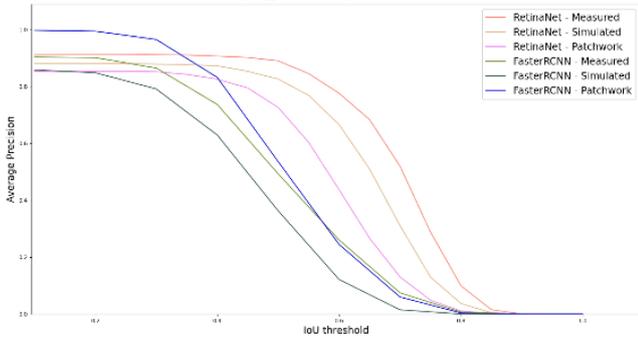

*Figure 12 - Average Precision over IoU threshold for Faster-RCNN and RetinaNet for the three different test datasets.*

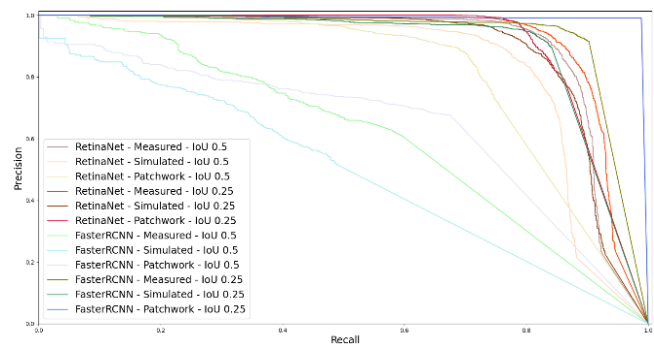

*Figure 13 - Precision x Recall curves of Faster-CNN and RetinaNet over the three test datasets and for 0.5 and 0.25 IoU threshold.*

## IV. CONCLUSION

In this work, we proposed a Deep Learning approach to train ATD models with synthetic SAR images produced with the MOCEM simulator. Our approach relies (1) on Deep Learning techniques to enhance the generalization capabilities of the models, and (2) on massive physic-based data-augmentation performed on-the-fly during the training to increase the representativeness of synthetic datasets.

To compensate for the lack of available data, we build our own test datasets using MSTAR measurements. By overlaying measured targets into measured backgrounds, we showed that our ATD models don't overfit the training simulated targets. By (1) using patchworks of measured MSTAR vignettes, and (2) overlaying simulated background objects (house and tree) in measured backgrounds, we showed that our ATD models do not exploit some potential overlaying bias to easily detect the object of interest. Thanks to our approach, ATD models trained on synthetic data reach an accuracy of almost 90 % on real measurements.

Our experiments on the different ATD datasets have demonstrated that our datasets are unbiased by the image overlay process. We have also shown that our ADASCA framework can be used to train SAR segmentation models with the synthetic data produced by MOCEM with good generalization on real measurements.

The data augmentation process may still be further improved thanks to the MOCEM simulator. Indeed, one could easily and efficiently introduce variability on EM materials as well as on the dimensions of the targets, and randomly modify the signature. We should in this case be able to (1) reproduce the variability of the real objects of interest met in operation, and (2) deal with the lack of information on these objects. To this end, we will take advantage of the EM scattering centers models (M3D-EM) produced by MOCEM [11]. Finally, future work could include the creation of synthetic background images in order to further improve the variability of ATD datasets.